\documentclass[runningheads]{llncs}
\usepackage{graphicx}
\usepackage{comment}
\usepackage{amsmath,amssymb} 
\usepackage{color}
\usepackage{booktabs}
\usepackage{cite}
\usepackage[belowskip=-15pt,aboveskip=5pt,font=small,labelfont=bf]{caption}
\usepackage[colorlinks,citecolor=blue]{hyperref}

\let\llncssubparagraph\subparagraph
\let\subparagraph\paragraph
\usepackage{titlesec}
\let\subparagraph\llncssubparagraph

\titlespacing*{\section}
{0pt}{4ex}{1ex}
\titlespacing*{\subsection}
{0pt}{2.5ex}{0.8ex}
\usepackage[width=122mm,left=12mm,paperwidth=146mm,height=193mm,top=12mm,paperheight=217mm]{geometry}

\begin{document}
\pagestyle{headings}
\mainmatter

\title{Semantic Curiosity for Active Visual Learning} 


\titlerunning{Semantic Curiosity for Active Visual Learning}
%
\author{Devendra Singh Chaplot\thanks{Equal Contribution}\inst{1} \and
Helen Jiang$^\star$\inst{1} \and
Saurabh Gupta\inst{2}\and 
Abhinav Gupta\inst{1}}
\authorrunning{Chaplot et al.}
%
\institute{Carnegie Mellon University \\
\email{\{chaplot,helenjia,abhinavg\}@cs.cmu.edu}\\
\and
UIUC \\
\email{saurabhg@illinois.edu}}
\maketitle

\vspace{-10pt}

\begin{center}
{\scriptsize Webpage: \url{https://devendrachaplot.github.io/projects/SemanticCuriosity}}
\end{center}
\vspace{-15pt}

\begin{abstract}
In this paper, we study the task of embodied interactive learning for object detection. Given a set of environments (and some labeling budget), our goal is to learn an object detector by having an agent select what data to obtain labels for. How should an exploration policy decide which trajectory should be labeled? One possibility is to use a trained object detector's failure cases as an external reward. However, this will require labeling millions of frames required for training RL policies, which is infeasible. Instead, we explore a self-supervised approach for training our exploration policy by introducing a notion of semantic curiosity. Our semantic curiosity policy is based on a simple observation -- the detection outputs should be consistent. Therefore, our semantic curiosity rewards trajectories with inconsistent labeling behavior and encourages the exploration policy to explore such areas. The exploration policy trained via semantic curiosity generalizes to novel scenes and helps train an object detector that outperforms baselines trained with other possible alternatives such as random exploration, prediction-error curiosity, and coverage-maximizing exploration.

\end{abstract}

\section{Introduction}
Imagine an agent whose goal is to learn how to detect and categorize objects. How should the agent learn this task? In the case of humans (especially babies), learning is quite interactive in nature. We have the knowledge of what we know and what we don't, and we use that knowledge to guide our future experiences/supervision. Compare this to how current algorithms learn -- we create datasets of random images from the internet and label them, followed by model learning. The model has no control over what data and what supervision it gets -- it is resigned to the static biased dataset of internet images. Why does current learning look quite different from how humans learn? During the early 2000s, as data-driven approaches started to gain acceptance, the computer vision community struggled with comparisons and knowing which approaches work and which don't. As a consequence, the community introduced several benchmarks from BSDS~\cite{MartinFTM01} to VOC~\cite{pascal-voc-2007}. However, a negative side effect of these benchmarks was the use of static training and test datasets. While the pioneering works in computer vision focused on active vision and interactive learning, most of the work in the last two decades focuses on static internet vision. But as things start to work on the model side, we believe it is critical to look at the big picture again and return our focus to an embodied and interactive learning setup.

\begin{figure}[t]
    \centering
    \includegraphics[width=1\linewidth]{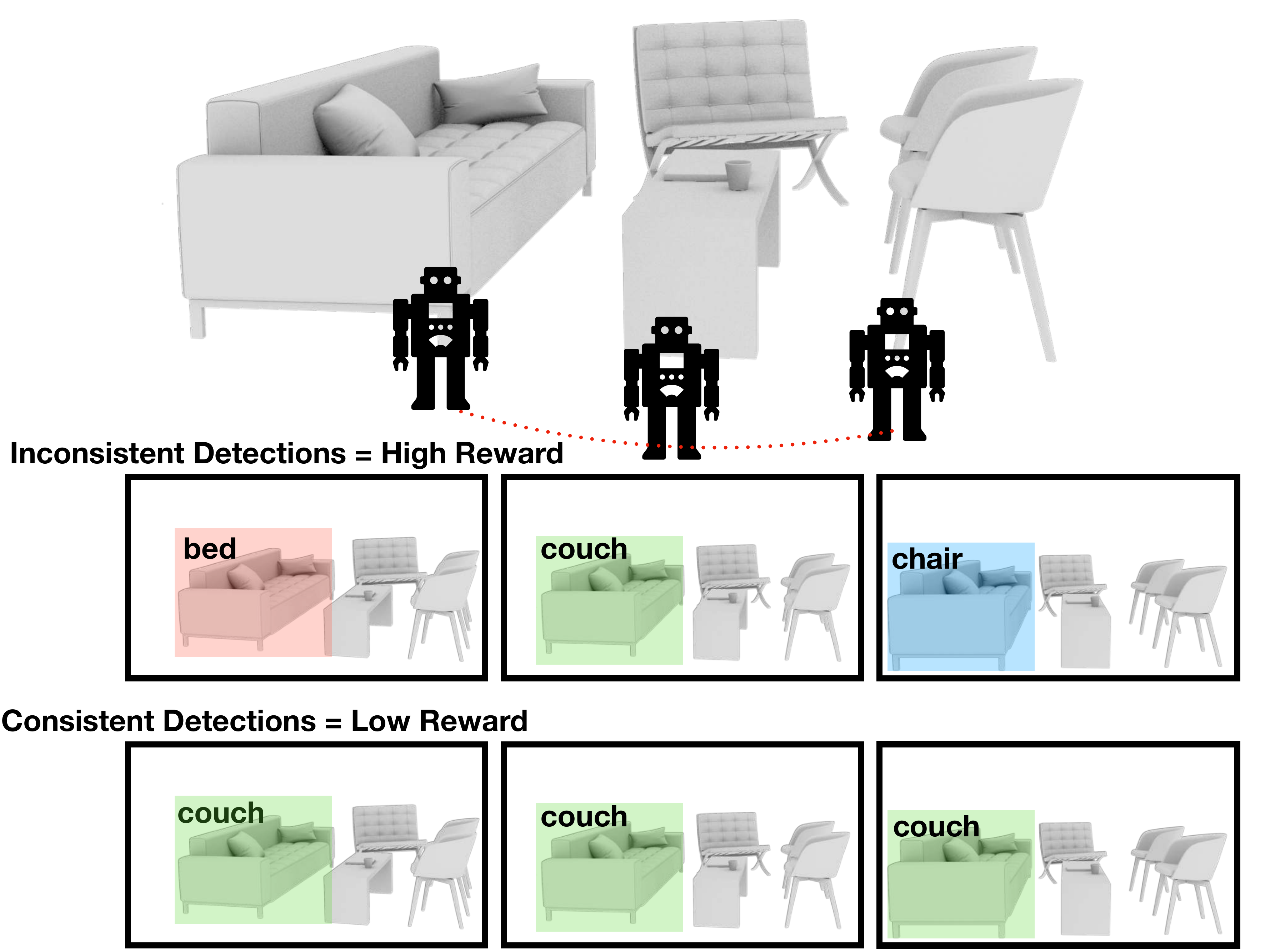}
    \caption{\textbf{Semantic Curiosity}: We propose semantic curiosity to learn exploration for training object detectors. Our semantically curious policy attempts to take actions such that the object detector will produce inconsistent outputs.}
    \label{fig:my_label}
\end{figure}

In an embodied interactive learning setup, an agent has to perform actions such that observations generated from these actions can be useful for learning to perform the semantic task. Several core research questions need to be answered: (a) what is the policy of exploration that generates these observations? (b) what should be labeled in these observations - one object, one frame, or the whole trajectory? (c) and finally, how do we get these labels? In this paper, we focus on the first task -- what should the exploration policy be to generate observations which can be useful in training an object detector? Instead of using labels, we focus on learning these trajectories in an unsupervised/self-supervised manner. Once the policy has been learned, we use the policy in novel (previously unseen) scenes to perform actions. As observations are generated, we assume that an oracle will densely label all the objects of interest in the trajectories. 

So what are the characteristics of a good exploration policy for visual learning, and how do we learn it? A good semantic exploration policy is one which generates observations of objects and not free-space or the wall/ceiling. But not only should the observations be objects, but a good exploration policy should also observe many unique objects. Finally, a good exploration policy will move to parts of the observation space where the current object detection model fails or does not work. Given these characteristics, how should we define a reward function that could be used to learn this exploration policy? Note, as one of the primary requirements, we assume the policy is learned in a self-supervised manner -- that is, we do not have ground-truth objects labeled which can help us figure out where the detections work or fail. 

Inspired by recent work in intrinsic motivation and curiosity for training policies without external rewards~\cite{pathak2017curiosity,pathak19disagreement}, we propose a new intrinsic reward called semantic curiosity that can be used for the exploration and training of semantic object detectors. In the standard curiosity reward, a policy is rewarded if the predicted future observation does not match the true future observation. The loss is generally formulated in the pixel-based feature space. A corresponding reward function for semantic exploration would be to compare semantic predictions with the current model and then confirm with ground-truth labels -- however, this requires external labels (and hence is not self-supervised anymore). Instead, we formulate semantic curiosity based on the meta-supervisory signal of consistency in semantic prediction -- that is, if our model truly understands the object, it should predict the same label for the object even as we move around and change viewpoints. Therefore, we exploit consistency in label prediction to reward our policies. Our semantic curiosity rewards trajectories which lead to inconsistent labeling behavior of the same object by the semantic classifier. Our experiments indicate that training an exploration policy via semantic curiosity generalizes to novel scenes and helps train an object detector which outperforms baselines trained with other possible alternatives such as random exploration, pixel curiosity, and free space/map curiosity. We also perform a large set of experiments to understand the behavior of a policy trained with semantic curiosity.

\section{Related Work}
\label{sec:related}
We study the problem of how to sample training data in embodied contexts. This is related to active learning (picking what sample to label), active perception (how to move around to gain more information), intrinsic motivation (picking what parts of the environment to explore). Learning in embodied contexts can also leverage spatio-temporal consistency. We survey these research areas below. 

\textbf{Active Perception.} Active perception~\cite{bajcsy1988active} refers to the problem of actively moving the sensors around at \textit{test time} to improve performance on the task by gaining more information about the environment. This has been instantiated in the context of object detection~\cite{ammirato2017dataset}, amodal object detection~\cite{yang2019embodied}, scene completion~\cite{jayaraman2018learning}, and localization~\cite{chaplot2018active, fox1998active}. We consider the problem in a different setting and study how to efficiently move around to best \textit{learn} a model. Furthermore, our approach to learn this movement policy is self-supervised and does not rely on end-task rewards, which were used in~\cite{jayaraman2018learning, chaplot2018active, yang2019embodied}. 

\textbf{Active Learning.} Our problem is more related to that of active learning~\cite{settles2009active}, where an agent actively acquires labels for unlabeled data to improve its model at the fastest rate~\cite{sener2017active, gal2017deep, yoo2019learning}. This has been used in a number of applications such as medical image analysis~\cite{kuo2018cost}, training object detectors~\cite{vijayanarasimhan2014large, yang2018visual}, video segmentation~\cite{fathi2011combining}, and visual question answering~\cite{misra2017lba}. Most works tackle the setting in which the unlabeled data has already been collected. In contrast, we study learning a policy for efficiently acquiring effective unlabeled data, which is complementary to such active learning efforts.

\textbf{Intrinsic Rewards.}
Our work is also related to work on exploration in reinforcement learning~\cite{schmidhuber1991possibility, sutton1998rli, auer2002using, jaksch2010near}. The goal of these works is to effectively explore a Markov Decision Process to find high reward paths. A number of works formulate this as a problem of maximizing an intrinsic reward function which is designed to incentivize the agent to seek previously unseen~\cite{eysenbach2018diversity} or poorly understood~\cite{pathak2017curiosity} parts of the environment. This is similar to our work, as we also seek poorly understood parts of the environment. However, we measure this understanding via multi-view consistency in semantics. This is in a departure from existing works that measure it in 2D image space~\cite{pathak2017curiosity}, or consistency among multiple models~\cite{pathak19disagreement}. Furthermore, our focus is not effective exhaustive exploration, but exploration for the purpose of improving semantic models.

\textbf{Spatio-Temporal smoothing at test time.}
A number of papers use spatio-temporal consistency at test time for better and more consistent predictions~\cite{chandra2018deep, gadde2017semantic}. Much like the distinction from active perception, our focus is using it to generate better data at train time.

\textbf{Spatio-temporal consistency as training signal.} 
Labels have been propagated in videos to simplify annotations~\cite{vatic}, improve prediction performance given limited data~\cite{badrinarayanan2010label, bengio200611}, as well as collect images~\cite{chen2013neil}. 
This line of work leverages spatio-temporal consistency to propagate labels for more efficient labeling. 
Researchers have also used multi-view consistency to learn about 3D shape from weak supervision~\cite{drcTulsiani19}. 
We instead leverage spatio-temporal consistency as a cue to identify what the model does not know. \cite{siddiqui2019viewal} is more directly related, but we tackle the problem in an embodied context and study how to navigate to gather the data, rather than analyzing passive datasets for what labels to acquire.

\textbf{Visual Navigation and Exploration.} Prior work on visual navigation can broadly be categorized into two classes based on whether the location of the goal is known or unknown. Navigation scenarios, where the location of the goal is known, include the most common \textit{pointgoal} task where the coordinate to the goal is given~\cite{gupta2017cognitive, savva2017minos}. Another example of a task in this category is vision and language navigation~\cite{anderson2018vision} where the path to the goal is described in natural language. Tasks in this category do not require exhaustive exploration of the environment as the location of the goal is known explicitly (coordinates) or implicitly (path). 

Navigation scenarios, where the location of the goal is not known, include a wide variety of tasks. These include navigating to a fixed set of objects~\cite{lample2016playing, dosovitskiy2016learning, wu2016training, chaplot2017arnold, mirowski2016learning, gupta2017cognitive}, navigating to an object specified by language~\cite{hermann2017grounded, chaplot2017gated} or by an image~\cite{zhu2017target, chaplot2020neural}, and navigating to a set of objects in order to answer a question~\cite{das2017embodied, gordon2018iqa}. Tasks in this second category essentially involve efficiently and exhaustively exploring the environment to search the desired object. However, most of the above approaches overlook the exploration problem by spawning the target a few steps away from the goal and instead focus on other challenges. For example, models for playing FPS games~\cite{lample2016playing, dosovitskiy2016learning, wu2016training, chaplot2017arnold} show that end-to-end RL policies are effective at reactive navigation and short-term planning such as avoiding obstacles and picking positive reward objects as they randomly appear in the environment. Other works show that learned policies are effective at tackling challenges such as perception (in recognizing the visual goal)~\cite{zhu2017target, chaplot2020neural}, grounding (in understanding the goal described by language)~\cite{hermann2017grounded, chaplot2017gated} or reasoning (about visual properties of the target object)~\cite{das2017embodied, gordon2018iqa}. While end-to-end reinforcement learning is shown to be effective in addressing these challenges, they are ineffective at exhaustive exploration and long-term planning in a large environment as the exploration search space increases exponentially as the distance to the goal increases. 

Some very recent works explicitly tackle the problem of exploration by training end-to-end RL policies maximizing the explored area~\cite{ans, chen2018learning, fang2019scene}. The difference between these approaches and our method is twofold: first, we train semantically-aware exploration policies as compared spatial coverage maximization in some prior works~\cite{ans, chen2018learning}, and second, we train our policy in an unsupervised fashion, without requiring any ground truth labels from the simulator as compared to prior works trained using rewards based on ground-truth labels~\cite{fang2019scene}.

\section{Overview}
Our goal is to learn an exploration policy such that if we use this policy to generate trajectories in a novel scene (and hence observations) and train the object detector from the trajectory data, it would provide a robust, high-performance detector. In literature, most approaches use on-policy exploration; that is, they use the external reward to train the policy itself. However, training an action policy to sample training data for object recognition requires labeling objects. Specifically, these approaches would use the current detector to predict objects and compare them to the ground-truth; they reward the policy if the predictions do not match the ground-truth (the policy is being rewarded to explore regions where the current object detection model fails). However, training such a policy via semantic supervision and external rewards would have a huge bottleneck of supervision. Given that our RL policies require millions of samples (in our case, we train using 10M samples), using ground-truth supervision is clearly not the way. What we need is an intrinsic motivation reward that can help train a policy which can help sample training data for object detectors.

\begin{figure*}[t]
    \centering
    \includegraphics[width=1\linewidth]{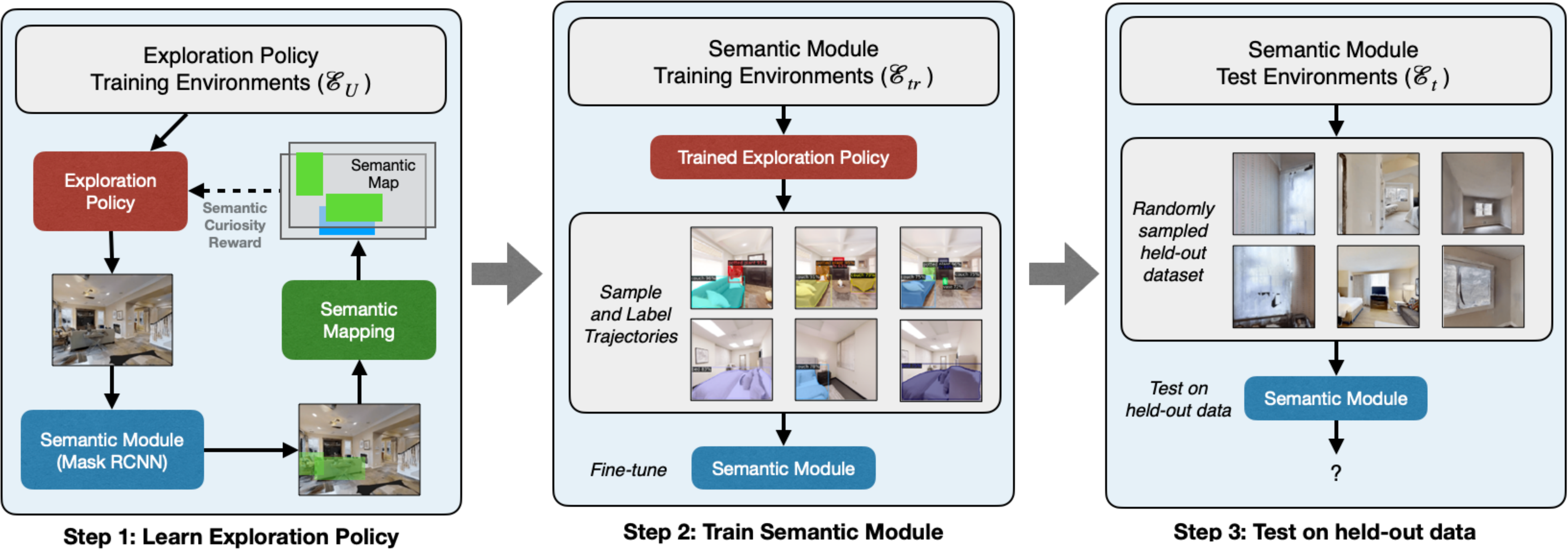}
    \caption{\textbf{Embodied Active Visual Learning}: We use semantic curiosity to learn an exploration policy on $\mathcal{E}_U$ scenes. The exploration policy is learned by projecting segmentation masks on the top-down view to create semantic maps. The entropy of semantic map defines the inconsistency of the object detection module. The learned exploration policy is then used to generate training data for the  object detection/segmentation module. The labeled data is then used to finetune and evaluate the object detection/segmentation.}
    \label{fig:overview}
\end{figure*}

We propose a semantic curiosity formulation. Our work is inspired by a plethora of efforts in active learning~\cite{settles2009active} and recent work on intrinsic reward using disagreement~\cite{pathak19disagreement}. The core idea is simple -- a good object detector has not only high mAP performance but is also consistent in predictions. That is, the detector should predict the same label for different views of the same object. We use this meta-signal of consistency to train our action policy by rewarding trajectories that expose inconsistencies in an object detector. We measure inconsistencies by measuring temporal entropy of prediction -- that is, if an object is labeled with different classes as the viewpoint changes, it will have high temporal entropy. The trajectories with high temporal entropy are then labeled via an oracle and used as the data to retrain the object detector (See Figure~\ref{fig:overview}). 

\section{Methodology}
Consider an agent $\mathcal{A}$ which can perform actions in environments $\mathcal{E}$. The agent has an exploration policy $a_t = \pi(x_t,\theta)$ that predicts the action that the agent must take for current observation $x_t$. $\theta$ represents the parameters of the policy that have to be learned. The agent also has an object detection model $\mathcal{O}$ which takes as input an image (the current observation) and generates a set of bounding boxes along with their categories and confidence scores. 

The goal is to learn an exploration policy $\pi$, which is used to sample $N$ trajectories $\tau_1...\tau_N$ in a set of novel environments (and get them semantically labeled). When used to train an object detector, this labeled data would yield a high-performance object detector. In our setup, we divide the environments into three non-overlapping sets $(\mathcal{E}_U, \mathcal{E}_{tr}, \mathcal{E}_t)$ -- the first set is the set of environments where the agent will learn the exploration policy $\pi$, the second set is the object detection training environments where we use $\pi$ to sample trajectories and label them, and the third set is the test environment where we sample random images and test the performance of the object detector on those images.

\begin{figure*}
    \centering
    \includegraphics[width=1\linewidth]{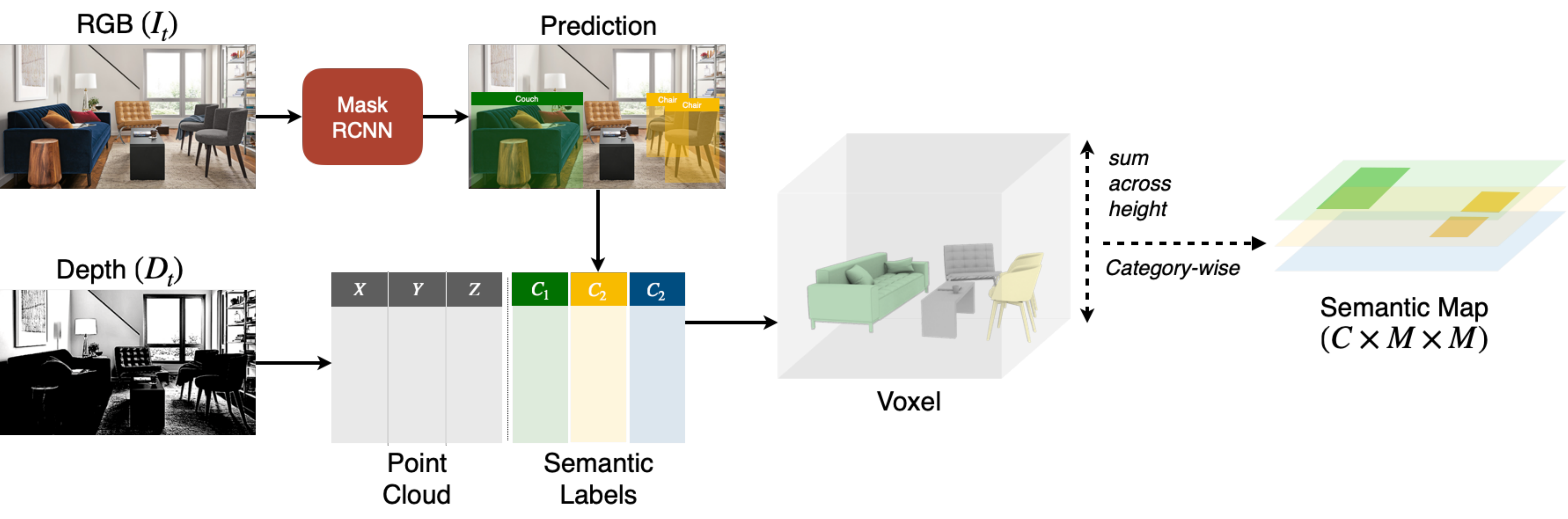}
    \caption{\textbf{Semantic Mapping.} The Semantic Mapping module takes in a sequence of RGB ($I_t$) and Depth ($D_t$) images and produces a top-down Semantic Map.}
    \label{fig:semantic mapping}
\end{figure*}

\subsection{Semantic Curiosity Policy}
\label{sec:semcur}
We define semantic curiosity as the temporal inconsistency in object detection and segmentation predictions from the current model. We use a Mask RCNN to obtain the object predictions. In order to associate the predictions across frames in a trajectory, we use a semantic mapping module as described below.\\

\noindent\textbf{Semantic Mapping.} The Semantic Mapping module takes in a sequence of RGB ($I_t$) and Depth ($D_t$) images and produces a top-down semantic map ($M^{Sem}_t$) represented by a 3-dimensional tensor $C \times M \times M$, where $M$ is the length of the square top-down map, and $C$ is the number of semantic categories. Each element $(c, i, j)$ in this semantic map is $1$ if the Mask RCNN predicted the object category $c$ at coordinates $(i,j)$ on the map in any frame during the whole trajectory and $0$ otherwise. Figure~\ref{fig:semantic mapping} shows how the semantic map is generated for a single frame. The input RGB frame ($I_t$) is passed through a current Mask RCNN to obtain object segmentation predictions, while the Depth frame is used to calculate the point cloud. Each point in the point cloud is associated with its semantic labels based on Mask RCNN predictions. Note that these are not ground-truth labels, as each pixel is assigned the category of the highest-confidence Mask RCNN segmentation prediction on the corresponding pixel. The point cloud with the associated semantic labels is projected to a 3-dimensional voxel map using geometric computations. The voxel representation is converted to a top-down map by max-pooling the values across the height. The resulting 2D map is converted to a 3-dimensional Semantic Map, such that each channel represents an object category.

The above gives a first-person egocentric projection of the semantic map at each time-step. The egocentric projections at each time step are used to compute a geocentric map over time using a spatial transformation technique similar to Chaplot et al.~\cite{ans}. The egocentric projections are converted to the geocentric projections by doing a spatial transformation based on the agent pose. The semantic map at each time step is computed by pooling the semantic map at the previous timestep with the current geocentric prediction. Please refer to~\cite{ans} for more details on these transformations. \\

\noindent\textbf{Semantic Curiosity Reward.} The semantic map allows us to associate the object predictions across different frames as the agent is moving. We define the semantic curiosity reward based on the temporal inconsistency of the object predictions. If an object is predicted to have different labels across different frames, multiple channels in the semantic map at the coordinates corresponding to the object will have $1$s. Such inconsistencies are beneficial for visual learning in downstream tasks, and hence, favorable for the semantic curiosity policy. Thus, we define the cumulative semantic curiosity reward to be proportional to the sum of all the elements in the semantic map. Consequently, the semantic curiosity reward per step is just the increase in the sum of all elements in the semantic map as compared to the previous time step:
$$
r_{SC} = \lambda_{SC} \Sigma_{(c,i,j) \in (C,M,M)} (M^{Sem}_t[c,i,j] - M^{Sem}_{t-1}[c,i,j])
$$
where $\lambda_{SC}$ is the semantic curiosity reward coefficient. Summation over the channels encourages exploring frames with temporally inconsistent predictions. Summation across the coordinates encourages exploring as many objects with temporally inconsistent predictions as possible. 

The proposed Semantic Curiosity Policy is trained using reinforcement learning to maximize the cumulative semantic curiosity reward. Note that although the depth image and agent pose are used to compute the semantic reward, we train the policy only on RGB images.

\section{Experimental Setup}
We use the Habitat simulator~\cite{savva2019habitat} with three different datasets for our experiments: Gibson~\cite{gibsonenv}, Matterport~\cite{Matterport3D} and Replica~\cite{replica19arxiv}.  While the RGB images used in our experiments are visually realistic as they are based on real-world reconstructions, we note that the agent pose and depth images in the simulator are noise-free unlike the real-world. Prior work has shown that both depth and agent pose can be estimated effectively from RGB images under noisy odometry~\cite{ans}. In this paper, we assume access to perfect agent pose and depth images, as these challenges are orthogonal to the focus of this paper. Furthermore, these assumptions are only required in the unsupervised pre-training phase for calculating the semantic curiosity reward and not at inference time when our trained semantic-curiosity policy (based only on RGB images) is used to gather exploration trajectories for training the object detector.\\

In a perfectly interactive learning setup, the current model's uncertainty will be used to sample a trajectory in a new scene, followed by labeling and updating the learned visual model (Mask-RCNN). However, due to the complexity of this online training mechanism, we show results on batch training. We use a pre-trained COCO Mask-RCNN as an initial model and train the exploration policy on that model. Once the exploration policy is trained, we collect trajectories in the training environments and then obtain the labels on these trajectories. The labeled examples are then used to fine-tune the Mask-RCNN detector.

\subsection{Implementation details}
\noindent {\bf Exploration Policy:} We train our semantic curiosity policy on the Gibson dataset and test it on the Matterport and Replica datasets. We train the policy on the set of 72 training scenes in the Gibson dataset specified by Savva et al.~\cite{savva2019habitat}. Our policy is trained with reinforcement learning using Proximal Policy Optimization~\cite{schulman2017proximal}. The policy architecture consists of convolutional layers of a pre-trained ResNet18 visual encoder, followed by two fully connected layers and a GRU layer leading to action distribution as well as value prediction. We use 72 parallel threads (one for each scene) with a time horizon on 100 steps and 36 mini batches per PPO epoch. The curiosity reward coefficient is set to $\lambda_{SC} = 2.5\times10^{-3}$. We use an entropy coefficient of 0.001, the value loss coefficient of 0.5. We use Adam optimizer with a learning rate of $1\times10^{-5}$.  The maximum episode length during training is 500 steps.

\noindent {\bf Fine-tuned Object Detector:} We consider 5 classes of objects, chosen because they overlap with the COCO dataset~\cite{lin2014microsoft} and correspond to objects commonly seen in an indoor scene: chair, bed, toilet, couch, and potted plant. To start, we pre-train a Faster-RCNN model~\cite{ren2015faster} with FPN~\cite{lin2017feature} using ResNet-50 as the backbone on the COCO dataset labeled with these 5 overlapping categories. We then fine-tuned our models on the trajectories collected by the exploration policies for 90000 iterations using a batch size of 12 and a learning rate of 0.001, with annealing by a factor of 0.1 at iterations 60000 and 80000. We use the Detectron2 codebase~\cite{wu2019detectron2} and set all other hyperparameters to their defaults in this codebase. We compute the AP50 score (i.e., average precision using an IoU threshold of 50) on the validation set every 5000 iterations. 

\begin{figure}[t]
    \centering
    \includegraphics[width=0.9\linewidth]{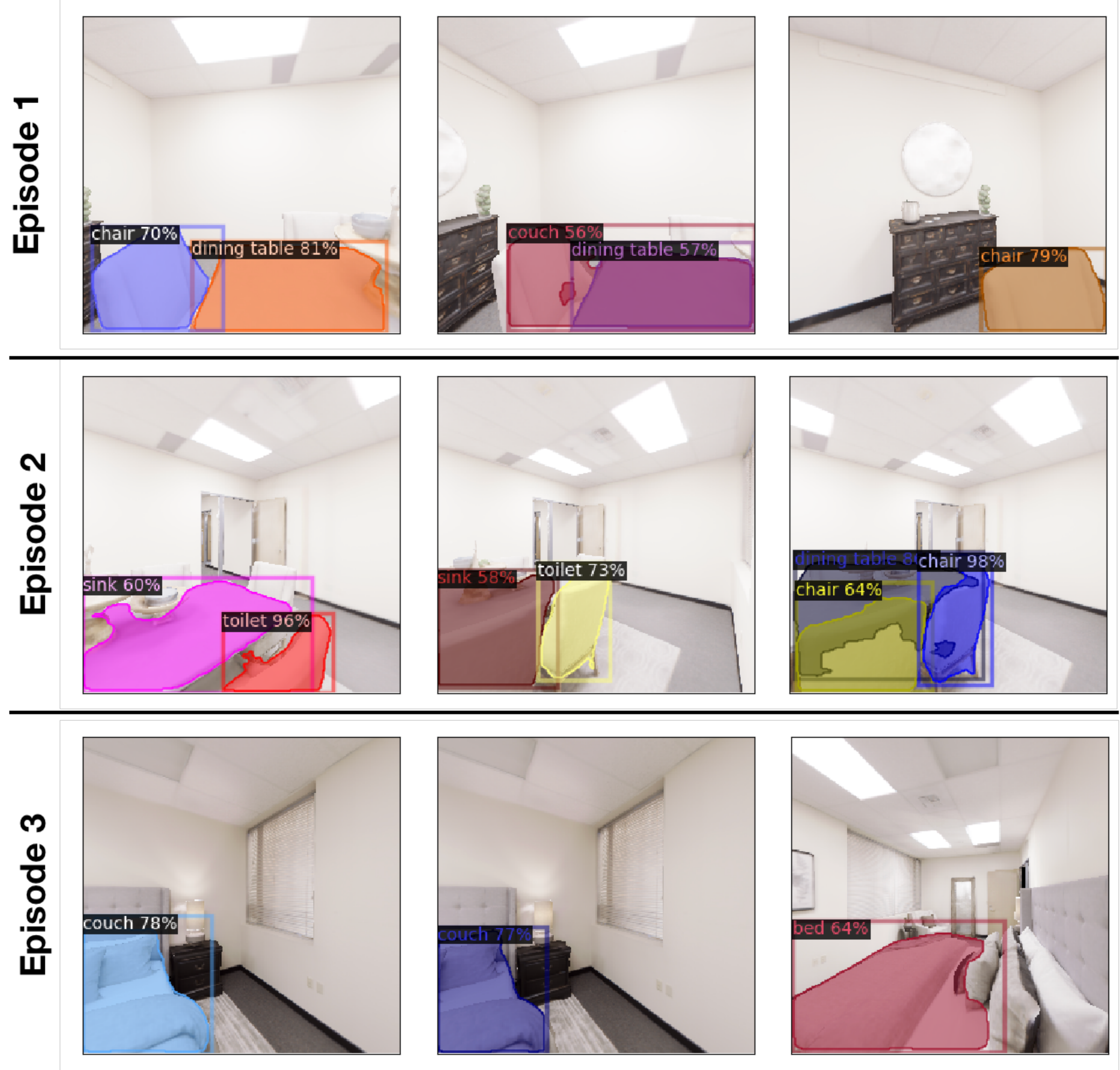}
    \caption{\textbf{Temporal Inconsistency Examples.} Figure showing example trajectories sampled from the semantic curiosity exploration policy. We highlight the segmentation/detection inconsistencies of Mask RCNN. By obtaining labels for these images, the Mask RCNN pipeline improves the detection performance significantly.}
  
    \label{fig:inconsistent}
\end{figure}

\subsection{Baselines}
We use a range of baselines to gather exploration trajectories and compare them to the proposed Semantic Curiosity policy:

\begin{itemize}
\item \textbf{Random.} A baseline sampling actions randomly.
\item \textbf{Prediction Error Curiosity.} This baseline is based on Pathak et al.~\cite{pathak2017curiosity}, which trains an RL policy to maximize error in a forward prediction model.
\item \textbf{Object Exploration.} Object Exploration is a naive baseline where an RL policy is trained to maximize the number of pre-trained Mask R-CNN detections. The limitation of simply maximizing the number of detections is that the policy can learn to look at frames with more objects but might not learn to look at different objects across frames or objects with low confidence.
\item \textbf{Coverage Exploration.} This baseline is based on Chen et al.~\cite{chen2018learning}, where an RL policy is trained to maximize the total explored area.
\item \textbf{Active Neural SLAM.} This baseline is based on Chaplot et al.~\cite{ans} and uses a modular and hierarchical system to maximize the total explored area.
\end{itemize}

After training the proposed policy and the baselines in the Gibson domain, we use them directly (without fine-tuning) in the Matterport and Replica domains. We sample trajectories using each exploration policy, using the images and ground-truth labels to train an object detection model. 

\setlength{\tabcolsep}{6.5pt}
\begin{table}[]
\centering
{\small
\caption{\textbf{Analysis.} Comparing the proposed Semantic Curiosity policy with the baselines along different exploration metrics.}
\label{tab:analysis}
\begin{tabular}{@{}llccc@{}}
\toprule
Method Name               &  & \begin{tabular}[c]{@{}c@{}}Semantic Curiosity\\ Reward\end{tabular} & \begin{tabular}[c]{@{}c@{}}Explored \\ Area\end{tabular} & \begin{tabular}[c]{@{}c@{}}Number of Object \\ Detections\end{tabular} \\ \midrule
Random                    &  & 1.631                                                              & 4.794                                                    & 82.83                                                          \\
Curiosity~\cite{pathak2017curiosity}                 &  & 2.891                                                              & 6.781                                                    & 112.24                                                         \\
Object exploration reward &  & 2.168                                                              & 6.082                                                    & 382.27                                                         \\
Coverage Exploration~\cite{chen2018learning}      &  & 3.287                                                              & 10.025                                                   & 203.73                                                         \\
Active Neural SLAM~\cite{ans}        &  & 3.589                                                              & 11.527                                                   & 231.86                                                         \\
Semantic Curiosity        &  & 4.378                                                              & 9.726                                                    & 291.78                                                         \\ \bottomrule
\end{tabular}
}
\end{table}

\setlength{\tabcolsep}{5.8pt}
\begin{table*}[t]
\centering
\caption{\textbf{Quality of object detection on training trajectories.} We also analyze the training trajectories in terms of how well the pre-trained object detection model works on the trajectories. We want the exploration policy to sample hard data where the pre-trained object detector fails. Data on which the pre-trained model already works well would not be useful for fine-tuning. Thus, lower performance is better.}
\label{tab:analysis2}
\begin{tabular}{@{}lcccccc@{}}
\toprule
Method Name          & Chair     & Bed     & Toilet    & Couch    & Potted Plant   & Average  \\
\midrule
Random                   & 46.7     & 28.2     & 46.9        & 60.3     & 39.1 & 44.24 \\
Curiosity~\cite{pathak2017curiosity}  &  49.4   &    18.3 &  1.8  &  67.7    & 49.0  &  37.42 \\
Object Exploration   & 54.3   &   24.8  & 5.7    &   76.6   & 49.6 & 42.2         \\
Coverage Exploration~\cite{chen2018learning} &  48.5   & 23.1  & 69.2  & 66.3 & 48.0  & 51.02  \\
Active Neural SLAM~\cite{ans}   & 51.3     & 20.5     & 49.4      & 59.7    & 45.6 & 45.3  \\
Semantic Curiosity  & 51.6      & 14.6    & 14.2      & 65.2     & 50.4 & 39.2  \\\bottomrule   
\end{tabular}
\end{table*}

\section{Analyzing Learned Exploration Behavior}
Before we measure the quality of the learned exploration policy for the task of detection/segmentation, we first want to analyze the behavior of the learned policy. This will help characterize the quality of data that is gathered by the exploration policy. We will compare the learned exploration policy against the baselines described above. For all the experiments below, we trained our policy on Gibson scenes and collected statistics in 11 Replica scenes. 

Figure~\ref{fig:inconsistent} shows three sampled trajectories. The pre-trained Mask-RCNN detections are also shown in the observation images. Semantic curiosity prefers trajectories with inconsistent detections. For example, in the top row, the chair and couch detector fire on the same object. In the middle row, the chair is misclassified as a toilet and there is inconsistent labeling in the last trajectory. The bed is misclassified as a couch. By selecting these trajectories and obtaining their labels from an oracle, our approach learns to improve the object detection module.

\begin{figure*}[t]
    \centering
    \includegraphics[width=0.99\linewidth]{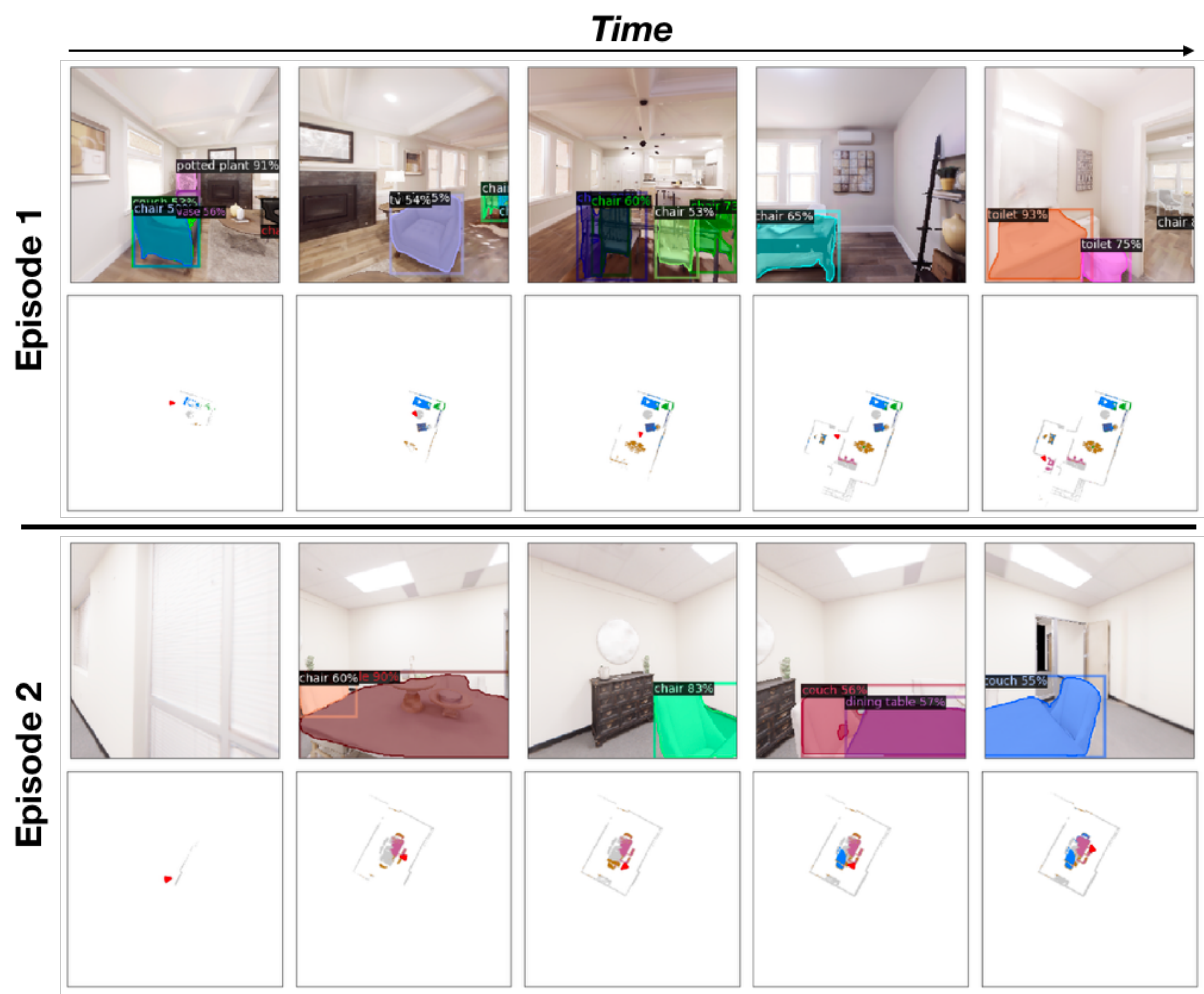}
    \caption{\textbf{Example trajectories.} Figure showing example trajectories sampled from the semantic curiosity exploration policy. In each episode the top row shows the first-person images seen by the agent and the pre-trained Mask R-CNN predictions. The bottom rows show a visualization of the semantic map where colors denote different object categories. Different colors for the same object indicate that the same object is predicted to have different categories from different view points.}
    \vspace{-0.1in}
    \label{fig:example_traj}
\end{figure*}

\begin{figure}[]
    \centering
    \includegraphics[width=0.99\linewidth]{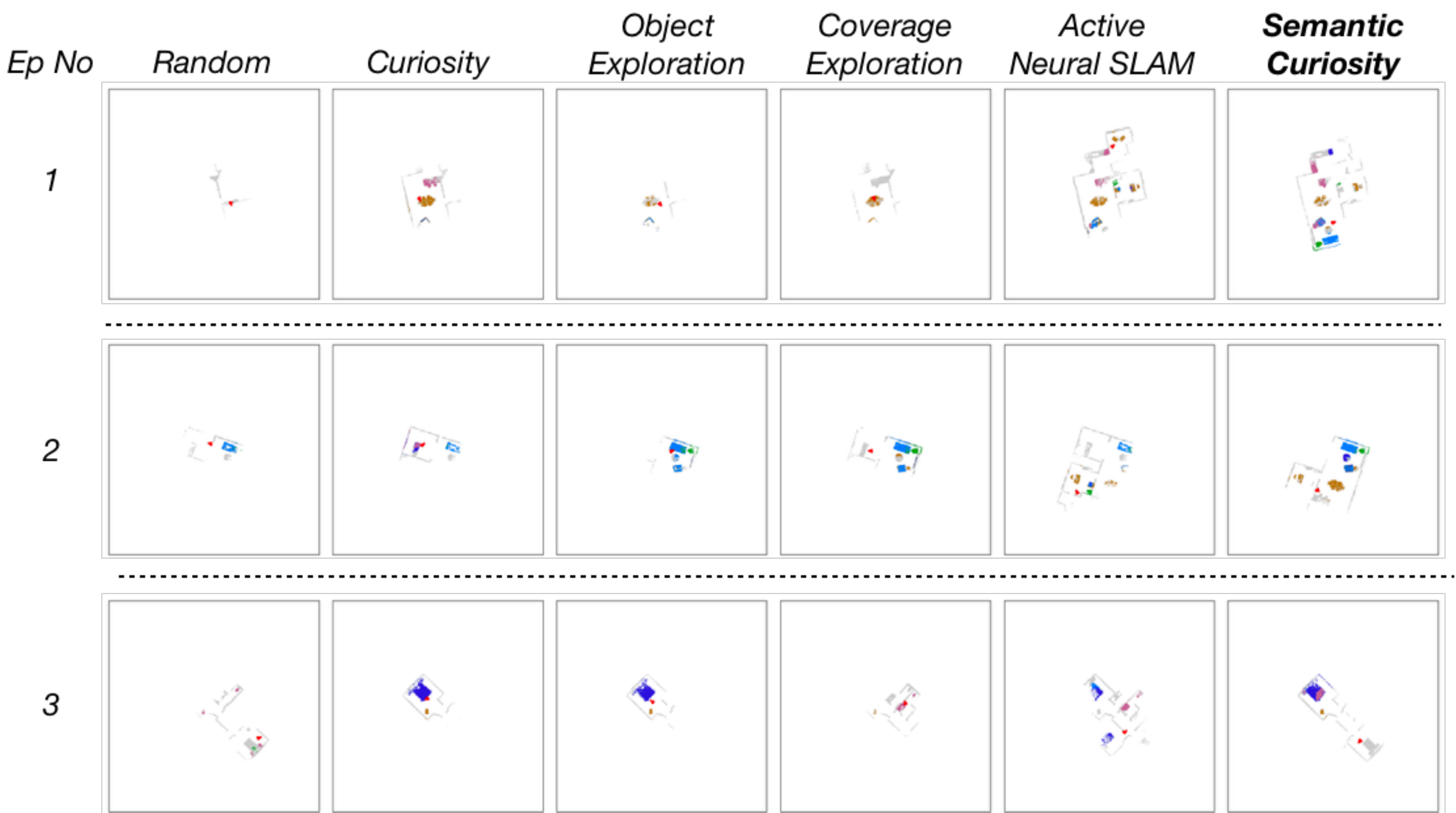}
    \caption{\textbf{Qualitative Comparison.} Figure showing map and objects explored by the proposed Semantic Curiosity policy and the baselines in 3 example episodes. Semantic Curiosity Policy explores more unique objects with higher temporal inconsistency (denoted by different colors for the same object).}
    \label{fig:Comparison}
\end{figure}

Table~\ref{tab:analysis} shows the behavior of all of the policies on three different metrics. The first metric is the semantic curiosity reward itself which measures uncertain detections in the trajectory data. Since our policy is trained for this reward, it gets the highest score on the sampled trajectories. The second metric is the amount of explored area. Both \cite{chen2018learning} and \cite{ans} optimize this metric and hence perform the best (they cover a lot of area but most of these areas will either not have objects or not enough contradictory overlapping detections). The third metric is the number of objects in the trajectories. The object exploration baseline optimizes for this reward and hence performs the best but it does so without exploring diverse areas or uncertain detections/segmentations. If we look at the three metrics together it is clear that our policy has the right tradeoff -- it explores a lot of area but still focuses on areas where objects can be detected. Not only does it find a large number of object detections, but our policy also prefers inconsistent object detections and segmentations. In Figure~\ref{fig:example_traj}, we show some examples of trajectories seen by the semantic curiosity exploration along with the semantic map. It shows examples of the same object having different object predictions from different viewpoints and also the representation in the semantic map. In Figure~\ref{fig:Comparison}, we show a qualitative comparison of maps and objects explored by the proposed model and all the baselines. Example trajectories in this figure indicate that the semantic curiosity policy explores more unique objects with higher temporal inconsistencies.

Next, we analyze the trajectories created by different exploration policies during the object detection training stage. Specifically, we want to analyze the kind of data that is sampled by these trajectories. How is the performance of a pre-trained detector on this data? If the pre-trained detector already works well on the sampled trajectories, we would not see much improvement in performance by fine-tuning with this data. In Table~\ref{tab:analysis2}, we show the results of this analysis for these trajectories. As the results indicate, the mAP50 score is low for the data obtained by the semantic curiosity policy.\footnote{Note that curiosity-based policy has the lowest mAP because of outlier toilet category.} As the pre-trained object detector fails more on the data sampled by semantic curiosity, labeling this data would intuitively improve the detection performance. 

\setlength{\tabcolsep}{5.8pt}
\begin{table}[t]
\centering
\caption{\textbf{Object Detection Results.} Object detection results in the Matterport domain using the proposed Semantic Curiosity policy and the baselines. We report AP50 scores on randomly sampled images in the test scenes. Training on data gathered from the semantic curiosity trajectories results in improved object detection scores.}
\label{tab:trained_ap}
\begin{tabular}{@{}lcccccc@{}}
\toprule
Method Name          & Chair     & Bed     & Toilet    & Couch    & Potted Plant   & Average  \\
\midrule
PreTrained               & 41.8      & 17.3     & 34.9      & 41.6     & 23.0    & 31.72   \\
Random                   & 51.7      & 17.2     & 43.0        & 45.1     & 30.0  & 37.4    \\
Curiosity~\cite{pathak2017curiosity}  & 48.4     &  18.5   &  42.3     &   44.3   &   32.8 &  37.26 \\
Object Exploration   & 50.3   & 16.4    &   40.0  &   39.7   & 29.9 &  35.26         \\
Coverage Exploration~\cite{chen2018learning} &  50.0    & 19.1    &   38.1    &  42.1   &  33.5 &  36.56      \\
Active Neural SLAM~\cite{ans}   & 53.1     & 19.5     & 42.0      & 44.5    & 33.4 & 38.5  \\
Semantic Curiosity   & 52.3      & 22.6    & 42.9      & 45.7     & 36.3 & \textbf{39.96}  \\\bottomrule   
\end{tabular}
\end{table}

\section{Actively Learned Object Detection}
Finally, we evaluate the performance of our semantic curiosity policy for the task of object detection. The semantic curiosity exploration policy is trained on 72 Gibson scenes. The exploration policy is then used to sample data on 50 Matterport scenes. Finally, the learned object detector is tested on 11 Matterport scenes. For each training scene, we sample 5 trajectories of 300 timesteps leading to 75,000 total training images with ground-truth labels. For test scenes, we randomly sample images from test scenes.

In Table~\ref{tab:trained_ap}, we report the top AP50 scores for each method. Our results demonstrate that the proposed semantic curiosity policy obtains higher quality data for performing object detection tasks over alternative methods of exploration. First, we outperform the policy that tries to see maximum coverage area (and hence the most novel images). Second, our approach also outperforms the policy that detects a lot of objects. Finally, apart from outperforming the random policy, visual curiosity~\cite{pathak2017curiosity}, and coverage; we also outperform the highly-tuned approach of ~\cite{ans}. The underlying algorithm is tuned on this data and was the winner of the RGB and RGBD challenge in Habitat.

\section{Conclusion and Future Work}
We argue that we should go from detection/segmentation driven by static datasets to a more embodied active learning setting. In this setting, an agent can move in the scene and create its own datapoints. An oracle labels these datapoints and helps the agent learn a better semantic object detector. This setting is closer to how humans learn to detect and recognize objects. In this paper, we focus on the exploration policy for sampling images to be labeled. We ask a basic question -- how should an agent explore to learn how to detect objects? Should the agent try to cover as many scenes as possible in the hopes of seeing more diverse examples, or should the agent focus on observing as many objects as possible? 

We propose semantic curiosity as a reward to train the exploration policy. Semantic curiosity encourages trajectories which will lead to inconsistent detection behavior from an object detector. Our experiments indicate that exploration driven by semantic curiosity shows all of the good characteristics of an exploration policy: uncertain/high entropy detections, attention to objects rather than the entire scene and also high coverage for diverse training data. We also show that an object detector trained on trajectories from a semantic curiosity policy leads to the best performance compared to a plethora of baselines. For future work, this paper is just the first step in embodied active visual learning. It assumes perfect odometry, localization and zero trajectory labeling costs. It also assumes that the trajectories will be labeled -- a topic of interest would be to sample trajectories with which minimal labels can learn the best detector. Finally, the current approach is demonstrated in simulators - it will be interesting to see whether the performance can transfer to real-world robots.

\section*{Acknowledgements}
\noindent This work was supported by IARPA DIVA D17PC00340, ONR MURI, ONR Grant N000141812861, ONR Young Investigator, DARPA MCS, and NSF Graduate Research Fellowship. We would also like to thank NVIDIA for GPU support.\\

{\scriptsize 
\noindent \textbf{Licenses for referenced datasets:}\\\vspace{-4pt}
Gibson: {\url{http://svl.stanford.edu/gibson2/assets/GDS_agreement.pdf}}\\\vspace{-4pt}
Matterport3D: {\url{http://kaldir.vc.in.tum.de/matterport/MP_TOS.pdf}}\\\vspace{-4pt}
Replica: {\url{https://github.com/facebookresearch/Replica-Dataset/blob/master/LICENSE}}
}
\clearpage
%
%
\bibliographystyle{splncs04}
\bibliography{egbib}
\end{document}